# Grounding Realizable Entities

Michael RABENBERG,[1,2*] Carter BENSON,[1,3] Federico DONATO,[1,3] Yongqun HE,[4] Anthony HUFFMAN,[4] Shane BABCOCK[1,5], John BEVERLEY[1,3,6]

[*]Corresponding Author

[1] National Center for Ontological Research, USA
[2] University at Buffalo Department of Biomedical informatics, USA
[3] University at Buffalo Department of Philosophy, USA
[4] University of Michigan Medical School, USA.
[5] KadSci LLC, USA
[6] Institute for Artificial Intelligence and Data Science, USA

**Abstract.** Ontological representations of qualities, dispositions, and roles have been refined over the past decade, clarifying subtle distinctions in life science research. After articulating a widely-used characterization of these entities within the context of Basic Formal Ontology (BFO), we identify gaps in this treatment and motivate the need for supplementing the BFO characterization. By way of supplement, we propose definitions for grounding relations holding between qualities and dispositions, and dispositions and roles, illustrating our proposal by representing subtle aspects of host-pathogen interactions.

## 1. Introduction

The prevalence of "roles" in contemporary life science literature is hard to overstate. Molecules are said to play roles in molecular transitions; enzymes play roles in metabolism; pacemakers play roles in heart rhythm regulation. Despite such prevalence, "role" is used in widely different ways across and within life science domains [1]. Some treat "role" as something a function can play, e.g. "role of mitochondria function" [2]. Others treat "role" as a type of function, e.g. "causal role function" [3]. Still others claim "role" refers to an ahistorical description of how an entity contributes to a complex system [4]. Varied uses of the same term runs the risk of creating confusion when there is a need for researchers to communicate across disciplines or indeed within a single discipline. The risk of confusion is exacerbated given the sheer volume of data across life science domains, more than can be mastered by a single individual in a lifetime. And while advances in computing, artificial intelligence (AI), machine learning, and so on, may hold the promise of automating detection of semantic differences and similarities in life science domains, training algorithms in training sets using the same label for semantically different phenomenon can only do so much [5]. Further advances must leverage, we maintain, knowledge representation methodologies in general, and ontology engineering in particular. Ontologies are widely used in bioinformatics and biomedical data standardization, supporting data integration, sharing, reproducibility, and automated reasoning [6]. Thus, in the interest of promoting clarity, AI-ready life science data, and semantic interoperability, we aim to improve on an ontological characterization of "role" and nearby phenomena.

More specifically, our goal in this piece is to cut through existing ambiguity in life science research and debates in ontological circles over the use of "roles" by supplementing what we take to be the best candidate interpretation of this term within the context of the widely-used Basic Formal Ontology (BFO) [7]. Despite the virtues of the BFO characterization, we argue the treatment of roles and related entities in BFO leave several important questions unanswered, and in doing so, obscures the - sometimes - subtle value offered by the BFO's definitions. To bring this value to light, we propose introducing grounding relations holding between roles and other dependent entities, such as dispositions and qualities. In defining precise ontological relationships among these terms, we will illustrate to stakeholders how they are intended to relate, providing guardrails for the use of both the terms and relationships among them. To illustrate the value of our proposal, we apply our results to one rather complicated modeling task: host-pathogen interactions. Our application demonstrates how appealing to grounding relations holding between properties can provide clarity when modeling even the most complex phenomena ontologists and researchers face.

## 1.1. *Basic Formal Ontology*

The Open Biomedical and Biological Ontologies (OBO) Foundry was created to provide guidance for ontology developers and promote alignment and interoperability while structuring life science data [7]. Basic Formal Ontology (BFO) has, for decades, been the designated OBO top-level ontology, providing life science researchers a highly general terms, such as **quality**,[1] **process**, **function**, and **role**, and relations, such as **participates in** and **member of**, which form a common starting point from which to extend to more specific terms and relations [7, 8]. At present count, BFO is used by over 600 open-source ontology projects [9], many of which aim to conform to OBO principles. Additionally, BFO is the first ISO/IEC approved top-level ontology standard 21838-2 [10], provides the foundation for the Industrial Ontologies Foundry [11], the Common Core Ontologies [12] suite, and is the "baseline standard" for ontology development across U.S. defense and intelligence agencies [13]. Given its wide use, standardization, and well-developed ontological structure, and existing characterization of roles and nearby phenomena, BFO is a natural starting point for exploring a more robust ontological characterization these entities.

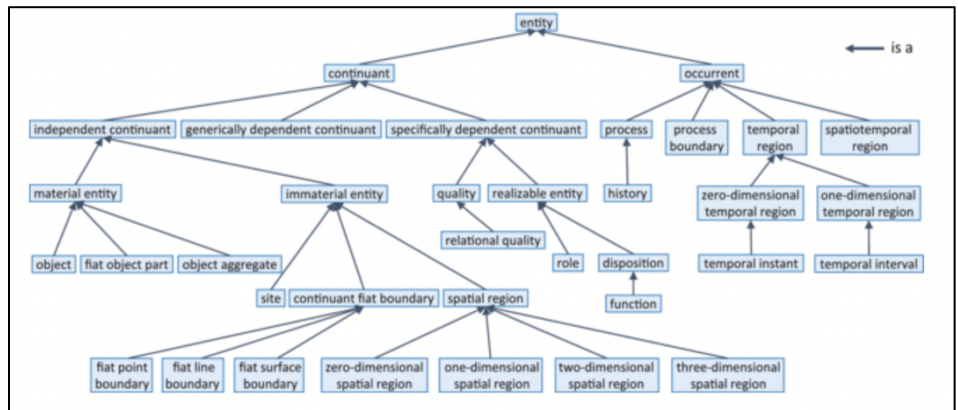

*Figure 1*: *Basic Formal Ontology Hierarchy*

---

[1]We adopt the convention of signifying BFO terms and relations in bold. Class names in BFO are always singular, i.e. "material entity" rather than "material entities". Plural class names in the text should be understood as shorthand for "instances of class X".

BFO divides reality into disjoint categories of **continuant** and **occurrent** [8]. **Continuants** lack temporal parts and exist entirely at any time at which they exist at all. **Occurrents**, in contrast, have temporal parts and so are often stretched along a temporal axis. BFO adopts several further subclasses of **continuant** and **occurrent**. *Figure 1* displays the complete BFO hierarchy where "A is_a B" means "instances of class A are instances of class B". This paraphrase, moreover, reflects BFO's distinguishing between instances, e.g. red on *this* apple, and classes, e.g. the *class* of red things. In this paper we focus on only those BFO entities relevant to the discussion of **role**.

*1.2.   Realizable Entities*

In BFO, all instances of **specifically dependent continuant**, including instances of **quality** and **realizable entity**, depend for their existence on other entities. Instances of **quality** are distinguished from those of **realizable entity**, however, in that while instances of the former are fully manifested whenever they exist, the latter are not. For example, a red apple bears a **quality**, an instance of the color red. This color will manifest in its entirety whenever it manifests at all. **Qualities** need not be of single bearers; the class **relational quality** covers, for example, instances that mutually depend on some entity, such as a relational quality of love or some manner of contractual relationship.

An apple plausibly also bears a tendency to decay if left in the heat, but this feature of the apple need not manifest for the apple to bear it. In BFO, what the apple also then bears, is an instance of **realizable entity** [14]. When investigating the natural world, identifying whether an apple *might* decay often involves attention to what similar types of entity *have done* in the past under similar environmental conditions, and prediction about what similar types of entity *will do* in the future. Instances of **realizable entity** can be understood as underwriting connections between what has happened and what may happen. Identifying instances of **realizable entity** thus requires attention to empirical evidence and inference to the best candidate explanatory mechanism. Instances of **realizable entity** always have some bearer on which they depend, which may participate in **processes** which realize the **realizable entity**. Instances of **realizable entity** may thus remain, in a sense, dormant. For example, a piece of sodium chloride need not ever dissolve if placed in unsaturated water. Instances of **realizable entity** are, in this respect, how BFO characterizes apparent modal and probabilistic features of reality.

There are further divisions to be drawn within the class. A student, for example, appears to bear a **realizable entity** insofar as they may or may not **participate in processes** associated with being a student. The sense in which a student bears a **realizable entity** is distinct from the sense in which salt is soluble. In the interest of capturing this distinction, BFO divides **realizable entity** into subclasses of **disposition** and **role**. The former is captured as [8]:[2]

> d is a **disposition** $=_{def}$  (i) d is a **realizable entity** & (ii) d's bearer is some **material entity** & (iii) d is such that, if it ceases to exist, then its bearer is physically changed & (iv) d's realization occurs because this bearer is in some special physical circumstances & (v) this realization occurs in virtue of the bearer's physical make-up.

---

[2]Talk of dispositions is also quite common in life science research. Brønsted-Lowry acids are said to be "disposed" to donate protons, infected patients are "disposed" to manifest symptoms of disease, and so on. Parallel to debates over "roles", there are persistent debates over "disposition," both in terms of ambiguity in the life science literature and extant debates among ontologists [15].

Instances of **disposition** are said to be "internally grounded", which reflects the fact that were the instance to cease to exist, then its bearer would be materially changed in some manner. For example, if a piece of sodium chloride is soluble at some time, then not soluble at another, there must be some change to its physical structure. In this sense, instances of **disposition** are not optional for bearers. Realizations of instances of **disposition** occur, moreover, owing to the physical structure of the bearer – determined by its **qualities** - and the fact that the bearer is in some special environment, where "special" means an environment the material bearer is not always in [8]. Turning again to the solubility of a piece of sodium chloride, the realization of its **disposition** occurs owing to the lattice structure and bonding forces of the **material entity** when placed in unsaturated water, which has its own bonding forces. *Figure 2* illustrates these features of sodium chloride solubility using the resources of BFO.

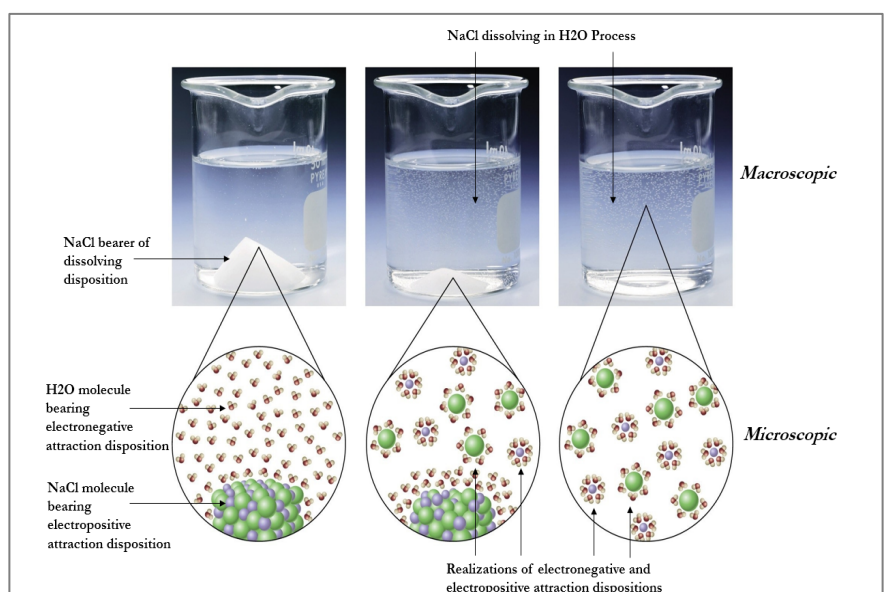

***Figure 2***: *A portion of NaCl manifesting its solubility disposition in a portion of $H_2O$*[3]

Role is a disjoint sibling class of **disposition**, and so a **realizable entity**, but one with characteristics sharply distinguishing it from its sibling [13]:[4]

r is a **role** =$_{def}$ (i) r is a **realizable entity** & (ii) r exists because there is some single bearer that is in some special physical, social, or institutional set of circumstances in which this bearer does not have to be & (iii) r is not such that, if it ceases to exist, then the physical make-up of the bearer is thereby changed.

---

[3]This diagram emphasizes one dimension of the chemical interaction, namely, the electronegativity of $H_2O$ and electropositivity of NaCl. A more detailed picture would have electropositive **dispositions** inhering in instances of Na+, and H+, and electronegative **dispositions** inhering in Cl- and O-.

[4]We note in passing that there remain debates among ontologists as to how understand "role" [16, 17].

In contrast to **disposition**, instances of **role** are thus optional in the sense that bearers may gain or lose them without alteration to their physical structure. A student, for example, who graduates from a university no longer bears the role of student at that institution. That does not, however, imply any physical change in the student. The individual, moreover, need not have been involved in that social circumstance at all. Put another way, whether an entity bears a role or not depends largely on what happens external to the entity. Related, **roles** may be borne by entities that do not have material matter as parts, i.e. **immaterial entities**, such as boundaries and sites. For example, the northern border between the U.S. and Canada plays a role in economic, immigration, and political policy for respective nations. The border itself is, however, not to be identified with any **material entity**, though its location may be determined by, say, geographical features between these countries. **Dispositions** cannot be borne by **immaterial entities** since **immaterial entities** do not have physical bases.

*1.3. The BFO Treatment*

The BFO characterization of **disposition** and its disjoint sibling class **role** tracks the following important distinctions (*inter alia*): (i) non-optional *vs.* optional; (ii) physical basis *vs.* non-physical basis; (iii) internally-grounded *vs.* externally-grounded. Call this the *BFO Treatment*. While the *BFO Treatment* provides classes and relations reflecting many aspects of these distinctions, there are noticeable gaps.

At a high level, little has been said about relationships among instances *within* each BFO class, such as the relationship between an employee role and a teacher role in a university [18] or the dispositional relationship between polar molecules [8]. Second, little has been said concerning relationships *across* classes, e.g. the relationship between a student role and the student's **disposition** to **participate in** coursework. Third, little has been said about relationships among instances – either within or across each class – at different *levels of granularity*, e.g. the relationship between the solubility of a molecule and the **qualities** born by its atomic structure [19]. This last point is worth belaboring as the **dispositions** a **material entity** bears are determined by some collection of **qualities** inhering in the bearer [14].[5] Although BFO adopts the **inheres in** relation holding between – among other things – instances of **quality** and instances of **material entity**, there is no corresponding relation holding between a given **disposition** and **qualities** in which it is grounded.

Ontology developers may be susceptible to – at best – ignoring these gaps or – at worst – modeling errors in specific domains. It is our position that supplementing the *BFO Treatment* with additional relationships among these classes, will help minimize user confusion and ontological errors. To sharpen our focus, we turn to explications of challenging modeling cases involving **roles** and related phenomena, to further motivate and highlight gaps in the *BFO Treatment*.

2. **Methods**

Development of the needed grounding relations proceeds by reflection on examples relevant to life science researchers. Careful analysis of entities which should be modeled ontologically in such cases, provides insight into features that these grounding relations must have. In general, such relationships may be quite complex. Realization of, say, the

---

[5] There is no "bare matter" which lacks qualities in BFO [20, 21].

**disposition** of sodium chloride to dissolve, affects the material structure of the system, and depends on realizations of dispositions borne by the other constituents.

> Case 1: Sodium chloride is soluble because of its lattice structure, a specific quality that it may lose at a time while remaining soluble, assuming the new structure it exhibits permits solubility. Sufficient changes to the lattice structure may result in sodium chloride no longer being soluble.

Properties may exhibit determinate-determinable structure [22, 23]. For example, an entity bearing color (determinable) may bear different colors over time (determinate). **Dispositions** also appear to exhibit determinate-determinable structure. Sodium chloride bearing a solubility disposition (determinable) may bear different degrees of solubility over time (determinate). Notably, in this example the determinable solubility appears to depend on determinate shapes, i.e. lattice structures, borne by the sodium chloride over time. Sufficient change in determinate shape may result in the loss of determinable or determinate solubility **dispositions**. As with other determinate properties, where cutoffs exist between one determinate **disposition** and another is an empirical question. That there may be no sharp cutoff is a feature of determinates [23]. Most importantly for our purposes, some collection of **qualities** makes it possible for a **disposition** to inhere in the bearer.

> Case 2: A university comprises, as mereological parts, administrative staff, educators, students, etc. each of which bear roles as members of the university, and these roles are correlated with various duties. A student, say, graduating from a university is no longer mereological part of the university. That is, a departing student results in material change of the university but may also lead to a change in the role the university has towards the student.

Universities can be understood as instances of BFO **object aggregate**, a **material entity** consisting solely of **object** member parts. Consequently, universities plausibly bear instances of **quality** and **realizable entity**, much like other **material entities**. Individuals gain and lose student roles – and so membership as part of a university – without material change to themselves. The loss of students, however, results in material change to a university. Sufficient loss of students, additionally, may even result in loss of **realizable entities** borne by the university itself, such as the **role** of a degree-granting institution. This suggests there are scenarios in which a university may lose a **role** while exhibiting change of physical basis through the loss of material parts. This parallels the dependence of **dispositions** on their physical basis,[6] suggesting a nuanced relationship between mereological parts of aggregates, roles, and dispositions that has as of present not been explored. From another direction, there is plausibly an instance of BFO's **relational quality** holding between the student and the university, which underwrites the **role** the former has. Once the individual is no longer a student, the relation between the individual and the university is eliminated. That is, there is a qualitative change to the university. Indeed, in any case of a student leaving university there will be such qualitative change.

---

[6] It is compatible with the *BFO Treatment* that entities bear both a **disposition** and a **role**. The heart of a gazelle might bear a **disposition** to pump blood, but in other circumstances may play the **role** of dinner for a carnivore.

> Case 3: In commensal relationships, like the one between a small baitfish swimming under the protection of a larger manta ray, one organism benefits while the other is neither benefitted nor harmed. If one organism leaves the relationship, it need not be physically changed, but the aggregate of the pair will lose a physical part.

This case is similar in structure to Case 2. But where students are mereological parts of a larger whole that exists after they depart, in this case the commensal aggregate no longer exists after a member departs. Each member of the pair bears a commensal role realized in commensal processes, which in our marine example is that of being protected and that of being a protector. The pair comprises an **object aggregate**, and each bears a respective commensal role insofar as they are members of the aggregate, which is easy to see when the baitfish departs from the manta ray and loses its protection, while no material part in its body or in that of the manta ray is necessarily changed. Though each may lose their respective role without being physically changed, the aggregate losing a mereological part results in a physical change, both in terms of matter and dispositions borne by the aggregate. As before, the grounds of each **role** might be understood in terms of **relational qualities** holding between the organisms. The organisms together comprise an **object aggregate** in which the **relational quality**, the relation of protecting, for example, inheres. Thus, when we describe an organism as bearing a commensal role, this is to say it bears a **realizable entity** grounded externally, in the sense that it is a proper part of some aggregate and other members of that aggregate partially ground the **realizable entity**.

These cases illustrate a variety of relationships involving **disposition**, **role**, and **quality** classes. Our characterizations of these cases highlight lines along which we may address gaps in the *BFO Treatment* of these entities. Case 1 illustrates the need for analyzing the relationship between **dispositions** and the **qualities** on which they depend. Case 2 and Case 3 highlight the need for clarifying relationships between aggregates, properties which they bear, properties borne by and among constituents. From our perspective, these cases illustrate a need for precise explication of varieties of grounding at play in these examples. We turn next to regimenting talk of grounding to supplement the *BFO Treatment*.

3. **Results**
Given the definitions of **role** and **disposition** in BFO, it seems clear that the latter can exist without the former. This, at least, seems intuitive when we restrict our attention to **material entity** bearers. A portion of water, for instance, bears many **dispositions**, but may not bear any **role** at a given time. That said, any given instance of **role** does seem to depend on some instance of **disposition**. It seems, moreover, that any instance of **disposition** must ultimately depend on some instance of **quality** or collection of **qualities**. This observation is motivated by the plausible assumption that entities bearing **qualities** can causally affect the world in some manner, which would be understood as the realization of some instance of **disposition**. Indeed, what would it mean to say some entity bears instances of **quality** but *no* instance of **disposition**? Much like we are hard-pressed to come up with an example of a bearer lacking all instances of **quality**, so too are we hard-pressed to come up with an example of a bearer lacking all instances of **disposition**.

*3.1.     Regimenting Grounding*
Case 1 illustrates how a **realizable entity** inhering in a **material entity** depends on the presence of a **quality** inhering in that **material entity**. More specifically:

> *x dependence grounded in y* $=_{def}$ *x* is a **realizable entity** and *y* is some **specifically dependent continuant** of type *T* and *x* **inheres in** *b* at *t* because determinates of *y* **inhere in** b at *t*.

Dependence grounding captures Case 1. For example, a piece of sodium chloride is an instance of **material entity** which bears an instance of **disposition**, which may manifest in dissolving **processes**. There is, moreover, some instance of a determinate shape **quality** inhering in the molecule of sodium chloride, on which the sodium chloride's **disposition** to dissolve depends. Were there no such instances of determinate shape **quality** of the relevant sort, then the portion of sodium chloride would not bear a **disposition** to dissolve.

Case 2 and Case 3 appear to exhibit a distinct variety of mereological dependence, in which the mereological parts of entities influence the presence or absence of realizable entities:

> *x mereologically grounded in y at $t_1$* $=_{def}$ *x* is a **realizable entity** inhering in **material entity** *y* with proper part *z* at a time $t_1$ and were *y* to become disjoint from *z* at some $t_2$ later than $t_1$ then *x* does not **inhere in** *y* at $t_2$

Mereological grounding applied to Case 2 makes sense of the fact that the loss of a sufficient number of students as mereological parts of the university may be accompanied by loss of some **role** inhering in the university. Similarly, mereological grounding makes sense of Case 3 as the commensal aggregate loses both **dispositions** and mereological parts following dissolution of the commensal relationship. We highlight **roles** given the lacuna in the literature, but it is worth observing mereological grounding can equally make sense of aggregates losing parts which bear instances of **disposition**, resulting in the aggregate losing a distinct instance of **disposition**. For example, a heart valve is disposed to expand and contract based on its elasticity, the loss of which, would undermine the heart of which it is mereological part, and which bears be a blood pumping **disposition**.

Mereological grounding can be explained in terms of dependence grounding. A university bearing a **disposition** that depends on maintaining a mereological part, bears some **specifically dependent continuant** on which this **disposition** depends. A pair of commensal organisms exhibit some manner of material structure in their pairing, which is a **specifically dependent continuant** on which **dispositions** borne by the pair depend. Loss of a mereological part often results in loss of a **specifically dependent continuant**, which in many cases brings loss of a **disposition** or **role** as well. Each of the three cases can be explained, it seems, by appealing to ultimately to dependence grounding alone.

Dependence grounding also provides explication of the *BFO Treatment* for **dispositions** as "internally grounded" and **roles** as "externally grounded". We maintain that in Case 1, the **disposition** of the portion of sodium chloride to dissolve is grounded in the **qualities** inhering in the **material entity**. The grounding of **roles**, in contrast, must refer to more than simply the bearer of the **role**, its **dispositions**, and **qualities**. A student bears a **role** that is grounded partially in the relevant person and partially in – for simplicity – the university. The aggregate itself bears **dispositions** which themselves depend on **qualities** of the aggregate, as discussed above. For example, the university is disposed to enroll new students each year, which depends on the quantity of students enrolled already and the financial resources of the university, among other things, many of which are in fact **relational qualities**. A student role is externally grounded in the university because it is grounded in instances of **relational quality** inhering in the university.

These remarks suggest the following: **dispositions** are internally grounded because they exhibit dependence grounding with respect to **qualities** that are not **relational qualities** while roles are externally grounded because they exhibit dependence grounding with respect to **relational qualities**. We thus have two species of dependence grounding:

> *x is internally grounded in y* $=_{def}$ *x* is a **realizable entity** and *y* is some **quality** of type *T* that is not a **relational quality** and *x* inheres in **material entity** *b* at a time *t* because determinates of *y* **inhere in** *b* at *t*.

> *x is externally grounded in y* $=_{def}$ *x* is a **realizable entity** and *y* is some **relational quality** of type *T* and *x* inheres in **material entity** *b* at a time *t* because determinates of *y* **inhere in** *b* at *t*.

The former applies to **dispositions**; the latter applies to **roles**.

3.2.  *Application to Host-Pathogen Interactions*

To illustrate how our proposal may aid researchers, we turn to the phenomena of host-pathogen interactions so important to contemporary infectious disease research. Until recently, microbiologists, immunologists, virologists, and others studying pathogenesis have engaged in either host-centered or pathogen-centered pathogenesis research. Each approach has its merits and has led to impressive research results. Nevertheless, emphasizing one aspect of host-pathogen interactions at the expense of the other may leave valuable questions unanswered. Emphasis, for example, solely on pathogenic factors of SARS-CoV-2 provide only a partial explanation of various pathogenesis pathways observed in clinical settings; focusing solely on host immune response is similarly limiting. These observations have not gone unnoticed by life science researchers and have motivated development of the Damage Response Framework [24] to explain host and pathogen contributions to pathogenesis. Defenders of the framework adopt the following principles, which guide our discussion:

(1) Pathogenesis results from interactions between host and pathogen and is attributable to neither alone.
(2) Host and pathogens interact primarily through damage to the host.
(3) Host damage is a function of the intensity and degree of host response and pathogen factors, each determined by genetic and phenotypic profiles.

Host and pathogen engage in – metaphorically – a tug of war, the results of which influence manifestations of signs, symptoms, and disease [25].

Even with such guidance we should take care to describe pathogens and hosts accurately, as these terms are used ambiguously in extant literature. For our purposes, "pathogen" should be understood as indexed either to a species or to stages in the developmental cycle of a species [26]. Motivation for the former stems from the fact that some viruses may engage in mutual symbiosis with one species, while exhibiting pathogenic behavior towards others. With respect to the latter, mature plants are often susceptible to different pathogens than developing plants. We reuse terms from the *Infectious Disease Ontology* for this illustration [26]:

> x is a **pathogenic disposition** =$_{def}$ **Disposition** borne by a **material entity** to establish localization in or produce toxins that can be transmitted to an organism or acellular structure, either of which may form disorder in the entity or immunocompetent members of the entity's species.

Pathogens bear such **dispositions** and so may establish localization in some **host**. Consider, *s. aureus* is an opportunistic pathogen in humans, becoming harmful to its **host** under changes in its environment. *S. aureus* is a **pathogen**, even when it does not realize disorder in a host, since it is nevertheless disposed to localize in a human host and generate disorder, if given the opportunity. This is a **disposition** of *s. aureus* – following BFO – because it is an "internally-grounded" property of the entity. That is, it is part of the material basis of *s. aureus* to generate disorder in human hosts if given the chance.

To characterize relationships between pathogens and hosts, IDO introduces:

> x is a **host role** =def **Role** borne by an acellular structure containing a distinct **material entity**, or organism whose extended organism contains a distinct **material entity,** realized in use of that structure or organism as a site of reproduction or replication.

**Hosts** thus bear externally-grounded **realizable entities**, namely, to be **sites** of reproduction or replication for **pathogens**. *Figure 3* illustrates various relationships emerging from host-pathogen interactions, using BFO conformant terminology and relations.

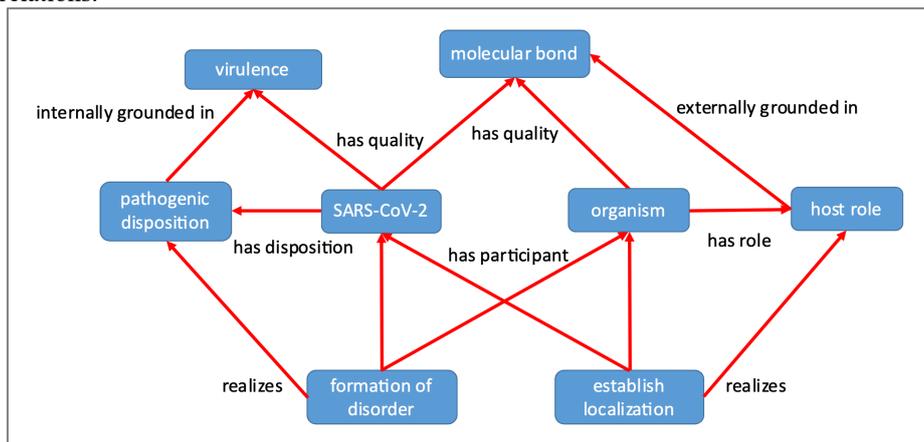

*Figure 3*: *Host Roles and Pathogen Dispositions*

An additional strength of our proposal is that we can reflect further subtleties arising from host-pathogen interactions. *Figure 5* illustrates the symmetry we find when focusing on the replication cycle of **pathogens** like SARS-CoV-2, which realizes an externally grounded pathogen role inhering in the virus. Similarly, host organisms bear host dispositions internally grounded in the quality of susceptibility inhering in the organism.

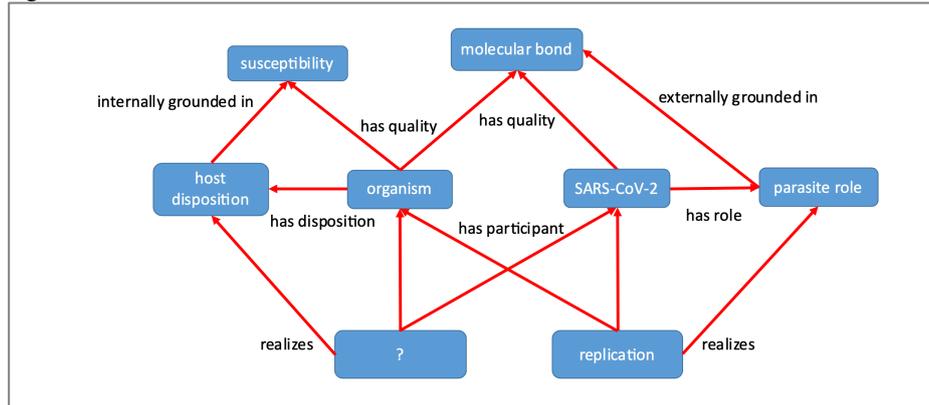

*Figure 4*: Host Dispositions and Pathogen Roles

4. **Discussion**

Our proposal clarifies relationships among **dispositions**, **roles**, and **qualities** within the *BFO Treatment*. We thus provide a supplement to ontologists using these terms to model such phenomena in various domains. Our proposal, moreover, provides precision to commonly used metaphors that make these terms obscure to those unfamiliar with the philosophical literature underwriting relevant definitions. **Roles** are **realizable entities** inhering in some member of an aggregate which bears a **relational quality** in which it is grounded; **dispositions** are **realizable entities** inhering in some entity which bears a **quality** or **qualities** that provide the grounds for the **disposition**.

There are of course limitations to our proposal worth noting. For example, implementing our proposal would require the inclusion of a new relation to BFO. BFO has undergone several changes in the past decade, moving from BFO 1.0 and 1.1 to BFO 2.0 to the present BFO-2020 specification. Presently, updates to BFO are handled locally. That is, ontologists importing versions of BFO are expected to conform to the most recent version, often requiring manual inspection and revision of domain-level ontological classes. Given the labor involved, version updates are not always completed quickly, and there is concern that users may experience "update fatigue" with new additions to downstream ontologies. We respect such concerns, but we note that while our supplement would require updating ontologies that import BFO, this update would require minimal effort. Our proposed changes do not change, for example, the hierarchy of BFO or alter existing relations associated with BFO. Our proposal is a simple additional resource which users can take advantage of when specifying complex relationships among **roles**, **dispositions**, and **qualities**.

5. **Concluding Remarks**

We have identified and defended a supplement to the *BFO Treatment* of **qualities**, **dispositions**, and **roles**. Gaps in this standard treatment can be filled by the introduction of grounding relations holding between **qualities** and **dispositions**, and **relational** **qualities** and **roles**. Among our goals in this proposal has been to lift any obscurity in the use of these entities among ontology developers and life science researchers. In the interest of showing developers how these relations can be applied, we provide ontological representations of subtle aspects of host-pathogen interactions.